\pdfoutput=1
\documentclass[conference,letterpaper]{IEEEtran}
\usepackage{times}
\usepackage{cite}
\usepackage{amsmath,amssymb}
\usepackage{graphicx}
\usepackage{booktabs}
\usepackage[bookmarks=true]{hyperref}
\hypersetup{hidelinks,
  pdftitle={Cut a Wall, See a Wall: Identity Substitution in Occluded Point-Cloud Perception for Construction Robots},
  pdfauthor={Tsung-Lung Yang},
  pdfsubject={Robot perception},
  pdfkeywords={point cloud segmentation, occlusion robustness, construction robotics, failure analysis, geometric features}}

\newcommand{\gle}{GLE$_\mathrm{pos}$}

\begin{document}

\title{\fontsize{17.28}{20.5}\selectfont\bfseries Cut a Wall, See a Wall: Identity Substitution in Occluded\\Point-Cloud Perception for Construction Robots}

\author{\IEEEauthorblockN{Tsung-Lung Yang}
\IEEEauthorblockA{National University\\
Email: webureetaipei@gmail.com}}

\maketitle

\begin{abstract}
Construction robots must label building elements they can only partly see. We show that when occlusion at test time removes whole regions of a scene, rather than parts of an object as in training, point-cloud segmentation networks do not simply miss the hidden elements: they relabel them, and the errors collapse onto a single \emph{sink} class. With fixed synthetic layouts the column absorbs up to 70\% of all test points; with randomized layouts the sink moves to the stair, predicted at 4.5--5.4$\times$ its true share across three training seeds. Baseline mIoU falls to 0.15--0.27 (fixed) and 0.36--0.72 (randomized), and up to 269 whole objects vanish in a single test set. Appending 14 hand-computed geometric features per point (GPE) restores mIoU to 0.54--0.92 and removes 64--100\% of whole-object losses in every run, for PointNet++ and a PTv3-style transformer in both layout regimes; on real S3DIS scans it removes every whole-wall loss (10$\to$0). In an as-built column check computed from the predictions, GPE raises the share of correctly checked views from 17--58\% to 71--92\%. Feature permutation on fixed, randomized and real scenes explains the pattern: each class relies on whichever cue is stable in its training data, position for height-banded slabs and beams, intrinsic shape for walls and stairs once their position varies. Occlusion augmentation is a strong alternative on mIoU, but in every setting it loses more whole objects than GPE (e.g., 25--35 vs.\ 0).
\end{abstract}


\section{Introduction}

Before a robot can use a laser scan of a building site, every point must be labelled: this point belongs to a column, that one to a beam. This semantic segmentation feeds localization, progress monitoring and safety checks \cite{fang2020,chern2023,guo2021}. On a real site the scan is never complete. Scaffolding, formwork, stacked material and the building itself hide part of every element, and which part is hidden changes as construction proceeds.

The common expectation is that occlusion hurts gradually: the more points are missing, the more often a hidden object is missed. We find that this expectation fails in a way that matters for a robot. When the occlusion at test time is of a kind the network never saw in training, it does not lose the object. It gives the object another identity, confidently and almost completely. In fixed synthetic layouts, removing the wall behind a column makes the network label the column \emph{wall}; removing the floor region makes the same column \emph{stair}. Across the test set, the errors pile up in one class that absorbs the others, while beams and stairs are erased. We call this \emph{identity substitution}.

\begin{figure*}[t]
\centering
\includegraphics[width=\textwidth]{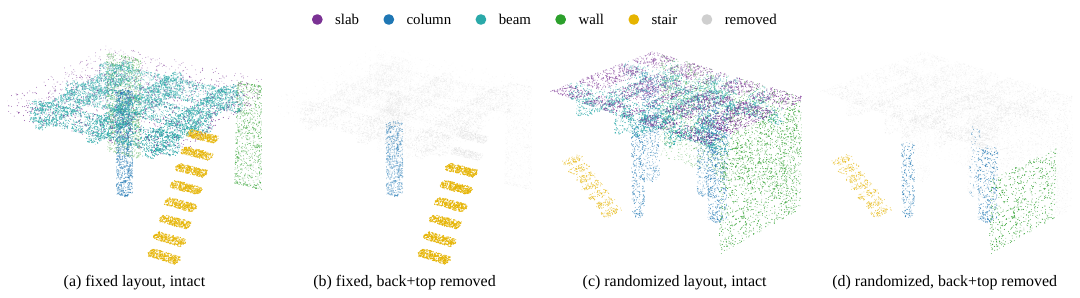}
\caption{The two scene families: (a) fixed layout, one column randomized; (c) randomized layout, everything drawn per scene. (b,d) An unseen scene-level cut removes the back and top of the whole scene (grey); in training, only the column itself was ever cut.}
\label{fig:scenes}
\end{figure*}

A natural objection is that this is an artifact of fixed layouts, where every element sits in the same place and position alone identifies it. We therefore rebuilt the benchmark with randomized room sizes, element counts, positions and orientations, and with density-neutral sampling. Substitution survives, but the sink moves from the column to the stair: its target is set by the training data, not by the occlusion alone.

This paper contributes the following.
\begin{itemize}
\item \textbf{A controlled measurement in two layout regimes.} Fixed and randomized synthetic rooms, seen and unseen occlusion splits generated from disjoint seeds, two architectures (PointNet++ and a PTv3-style transformer), three training seeds, and whole-object losses reported next to mIoU.
\item \textbf{Identity substitution with a data-dependent sink.} Under unseen scene-level occlusion the errors concentrate on one class: the column under fixed layouts, the stair under randomized ones.
\item \textbf{A cue-stability account of the mechanism.} Permutation importance on fixed, randomized and real (S3DIS) scenes shows that classes with a stable height keep relying on position, while walls and stairs switch to intrinsic shape once their position varies.
\item \textbf{A repair and an honest comparison.} Fourteen geometric features (GPE) cut whole-object losses by 64--100\% in every run, make an as-built column check correct in 71--92\% of views instead of 17--58\%, and transfer to real scans for walls. We compare them with occlusion augmentation, report where GPE fails, and show that only the tail-loss advantage, not the mIoU advantage, is stable across training seeds.
\end{itemize}

\section{Related Work}

\textbf{Point-cloud segmentation.} Point-based networks process unordered point sets directly, from PointNet and PointNet++ \cite{pointnet,pointnet2} through convolutional \cite{pointconv,kpconv,minkowski}, graph \cite{dgcnn}, MLP \cite{pointmlp} and transformer designs \cite{stratified,ptv2,ptv3}, with modernized training recipes \cite{pointnext}. We study two ends of this range, PointNet++ and a serialized-attention model in the spirit of PTv3, to separate architecture-specific behavior from mechanism.

\textbf{Robustness and occlusion.} Corruption benchmarks quantify how point cloud models degrade under noise, dropout and density change \cite{ren2022,robo3d,xiao2023}, and occlusion-specific benchmarks have appeared \cite{belhouse}. In construction and MEP scenes, occlusion-aware segmentation \cite{jing2025hybrid,meppnext} and completion \cite{pointr} are active topics. These works summarize robustness by how much accuracy is lost. We ask what the model outputs instead, and which input evidence drives it.

\textbf{Handcrafted geometric features.} Eigenvalue-based shape descriptors have a long history in point-cloud interpretation \cite{weinmann,hackel,matrone,kassotakis} and remain effective in data-scarce industrial settings \cite{noichl,yin2022,scan2bim}. We use them as an intervention: a controlled way to supply object-intrinsic evidence and test whether the failure is informational or architectural. We attribute reliance with permutation importance \cite{breiman}.

\section{Setup}

\textbf{Classes and scene families.} Every scene contains up to five structural classes $\{$slab, column, beam, wall, stair$\}$ (Fig.~\ref{fig:scenes}). \emph{Fixed layouts} keep the slab, a $3{\times}3$ grid of six beams, two walls and an eight-step stair in place and randomize the position and cross-section of one column. Because nothing else moves, every failure can be traced to a removed element. Each object receives a fixed number of points (1,200; 400 per stair step). \emph{Randomized layouts} draw the room size ($6$--$12$\,m $\times$ $6$--$12$\,m, height $2.8$--$3.6$\,m), 1--4 columns ($0.35$--$0.70$\,m), beams along column lines plus 0--2 extra beams, 1--4 wall segments on random sides, a stair with probability 0.8 along a random side, and a random yaw for the whole scene. Height relations that hold in real buildings (slab on top, beams under the slab) are kept. Points are sampled on visible surfaces at a uniform density per scene ($70$--$100$\,pts/m$^2$), so local density carries no class information; faces hidden by contact or facing out of the room are removed, and 3\,mm jitter is added. Views hold about 15k points before occlusion.

\textbf{Two kinds of occlusion.} This distinction is the core of the experiment. \emph{Training occlusion} removes a slice of one column along its own axes (top, bottom, left, right, a middle band, or two two-way combinations) at three severity levels; the rest of the room, and with it the scene's coordinate frame, is untouched. The \emph{seen} test set (test1, 430 views) is drawn from this family. \emph{Unseen occlusion} instead deletes whole regions of the scene: 25--45\% of the scene extent along each chosen direction, or a central box spanning 78--92\% of each axis for the \emph{middle} cut. test2 has 220 views over 11 combinations (nine two-direction, two three-direction); test3 has 160 views over eight three-direction combinations. A cut removes 18--67\% of a scene's points (about 68\% for the middle cut), and context classes disappear together with the target. Every view keeps at least 30 column points. Training, test1, test2 and test3 are generated from disjoint seeds, and we verify that no layout is shared between training and any test set.

\textbf{GPE.} The geometric positional encoding appends 14 handcrafted channels to each point, computed from $k{=}20$ neighborhoods with local covariance eigenvalues $e_1{\ge}e_2{\ge}e_3$ and normal $\mathbf{n}$: linearity, planarity and scattering (LSE, ch.~3--5); $x, y, z$ normalized to the scene box and height above the floor (\gle, 6--9); verticality $|\mathbf{n}{\cdot}\hat{\mathbf{z}}|$ (VERT, 10); mean and spread of neighbor distances and local density (DPE, 11--13); and normal consistency, plane residual and distance irregularity (OD, 14--16). Channels are standardized with training statistics. Channel 14 repeats the formula of channel 4, so OD and LSE partly share credit in the attribution of Sec.~\ref{sec:why}. Baselines receive coordinates only; GPE models receive 17 channels. Everything else is identical within each pair.

\textbf{Models and metrics.} PointNet++ follows the standard set abstraction/feature propagation design \cite{pointnet2}, trained on 12,000-point samples (20 epochs, learning rate $10^{-3}$, inverse-frequency class weights) and evaluated on the full scene in 12,000-point chunks. The second model is a simplified serialized transformer in the spirit of PTv3 \cite{ptv3}: single resolution, Z-order serialization, windowed attention (width 64, 4 heads, depth 4, window 256), trained on 4,096-point samples with 512 guaranteed column points (40 epochs) and evaluated on a random 4,096-point subsample per view. We call it PTv3-style. We report per-class IoU and mIoU from global confusion matrices and a \emph{whole-object loss}: a view in which a class with ${\geq}30$ ground-truth points has recall ${\leq}0.1$. Losses are compared with exact McNemar tests over views. Randomized-layout baselines and GPE models are trained three times (the original run and seeds 2 and 3); fixed-layout models once.

\section{What Unseen Occlusion Does}
\label{sec:what}

\begin{table}[t]
\centering
\caption{mIoU and whole-object losses, baseline$\to$+GPE. Seen = test1 (object cuts, as in training); test2/test3 = unseen scene-level cuts. Randomized layouts: mean of three training seeds (ranges in Sec.~\ref{sec:what}).}
\label{tab:main}
\footnotesize
\setlength{\tabcolsep}{3pt}
\begin{tabular}{llccccc}
\toprule
 & & \multicolumn{3}{c}{mIoU} & \multicolumn{2}{c}{objects lost}\\
\cmidrule(lr){3-5}\cmidrule(lr){6-7}
layout & model & seen & test2 & test3 & test2 & test3\\
\midrule
fixed & PN++ & .85$\to$.95 & .20$\to$.79 & .15$\to$.65 & 269$\to$3 & 202$\to$0\\
fixed & PTv3 & .96$\to$.98 & .27$\to$.67 & .17$\to$.54 & 107$\to$17 & 123$\to$16\\
\addlinespace
random & PN++ & .90$\to$.96 & .54$\to$.89 & .36$\to$.82 & 184$\to$17 & 237$\to$9\\
random & PTv3 & .95$\to$.98 & .71$\to$.92 & .55$\to$.90 & 21$\to$4 & 26$\to$9\\
\bottomrule
\end{tabular}
\end{table}

\begin{table}[t]
\centering
\caption{Whole-object losses per class, baseline$\to$+GPE (exact McNemar $p$ for column and wall). Randomized rows: first training seed.}
\label{tab:perclass}
\scriptsize
\setlength{\tabcolsep}{2.5pt}
\begin{tabular}{lcccccc}
\toprule
setting & test & slab & column ($p$) & beam & wall ($p$) & stair\\
\midrule
fixed PN++ & 2 & 3$\to$0 & 18$\to$1 (7.6e-5) & 109$\to$0 & 49$\to$0 (3.6e-15) & 90$\to$2\\
fixed PN++ & 3 & 22$\to$0 & 15$\to$0 (6.1e-5) & 66$\to$0 & 43$\to$0 (2.3e-13) & 56$\to$0\\
fixed PTv3 & 2 & 0$\to$0 & 48$\to$17 (7.9e-9) & 25$\to$0 & 25$\to$0 (6.0e-8) & 9$\to$0\\
fixed PTv3 & 3 & 7$\to$0 & 31$\to$13 (4.0e-5) & 25$\to$1 & 33$\to$1 (4.7e-10) & 27$\to$1\\
rand.\ PN++ & 2 & 0$\to$0 & 84$\to$22 (7.0e-18) & 9$\to$1 & 87$\to$0 (1.3e-26) & 1$\to$1\\
rand.\ PN++ & 3 & 0$\to$0 & 115$\to$9 (6.7e-31) & 3$\to$0 & 116$\to$0 (2.4e-35) & 2$\to$1\\
rand.\ PTv3 & 2 & 0$\to$0 & 13$\to$5 (.039) & 0$\to$0 & 6$\to$0 (.031) & 2$\to$0\\
rand.\ PTv3 & 3 & 0$\to$0 & 12$\to$5 (.065) & 0$\to$0 & 8$\to$0 (.0078) & 7$\to$4\\
\bottomrule
\end{tabular}
\end{table}

\begin{table}[t]
\centering
\caption{As-built column check from the predictions, baseline$\to$+GPE (\%). Missed: visible columns with $<$50\% of points labelled column. Alarm: views with a phantom vertical column member. Correct: views with neither. Randomized rows: mean over three seeds (range of correct for GPE).}
\label{tab:check}
\footnotesize
\setlength{\tabcolsep}{4pt}
\begin{tabular}{lcccccc}
\toprule
layout & model & test & missed & alarm & correct\\
\midrule
fixed & PN++ & 2 & 14$\to$2 & 44$\to$12 & 45$\to$88\\
fixed & PN++ & 3 & 11$\to$3 & 34$\to$17 & 58$\to$82\\
fixed & PTv3 & 3 & 38$\to$13 & 33$\to$17 & 42$\to$71\\
random & PN++ & 2 & 51$\to$9 & 20$\to$0 & 35$\to$88 (81--94)\\
random & PN++ & 3 & 82$\to$15 & 5$\to$0 & 17$\to$77 (63--90)\\
random & PTv3 & 2 & 21$\to$7 & 19$\to$0 & 56$\to$88 (87--89)\\
random & PTv3 & 3 & 33$\to$6 & 18$\to$0 & 43$\to$92 (89--95)\\
\bottomrule
\end{tabular}
\end{table}

\textbf{Seen occlusion is nearly harmless.} On test1 both baselines are strong in both regimes (mIoU 0.85--0.96, Table~\ref{tab:main}). The only class with whole-object losses is the column, and they are few: 13$\to$6 of 423 views for PointNet++ ($p{=}.016$) and 10$\to$3 of 386 for PTv3 ($p{=}.065$) in the fixed family, 9$\to$5 ($p{=}.22$) and 1$\to$1 in the randomized family (5--10$\to$4--5 and 1--2$\to$1 across the three seeds).

\begin{figure*}[t]
\centering
\includegraphics[width=\textwidth]{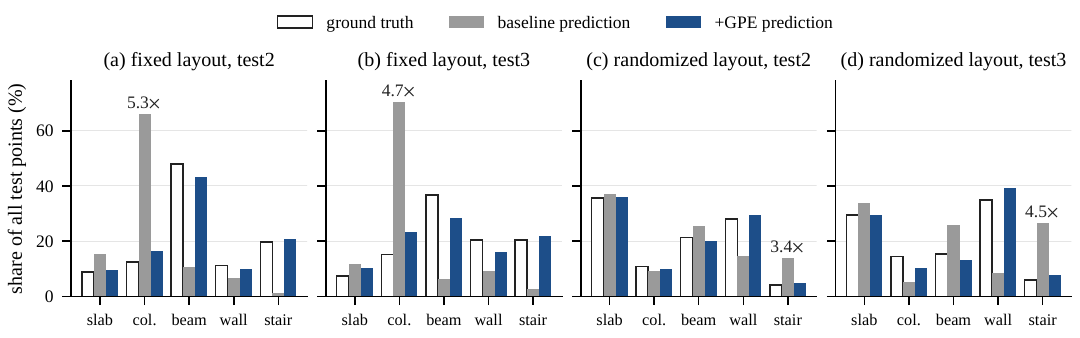}
\caption{Where the points go (PointNet++). Share of all test points per class: ground truth, baseline prediction and GPE prediction. The baseline over-predicts one sink class, the column in fixed layouts and the stair in randomized layouts; GPE restores the true shares. Annotations give the sink's over-prediction factor. Randomized panels: first training seed (the other two give 3.5--4.0$\times$ on test2 and 4.7--5.4$\times$ on test3).}
\label{fig:shares}
\end{figure*}

\begin{figure*}[t]
\centering
\includegraphics[width=\textwidth]{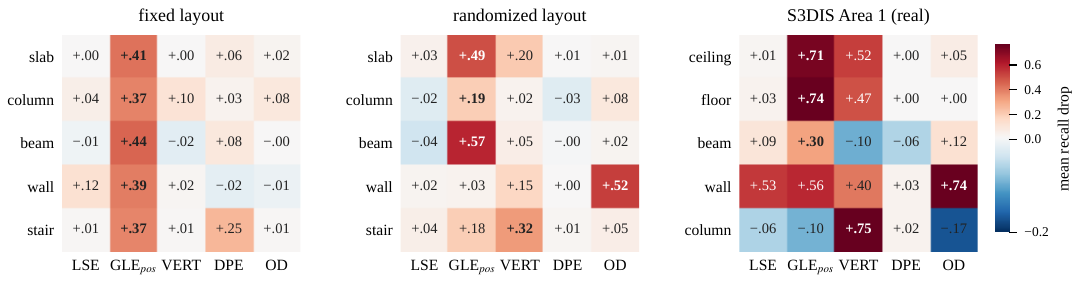}
\caption{Permutation importance of the PointNet++ GPE model: mean drop in per-view recall when one channel group is shuffled within a view (3 repeats, paired chunk order, all test2 views, first training seed; S3DIS: all 139 test views). Bold = the largest carrier of each class.}
\label{fig:perm}
\end{figure*}

\textbf{Unseen occlusion collapses the class system.} Under scene-level cuts baseline mIoU falls to 0.15--0.27 with fixed layouts and to 0.36--0.72 with randomized ones. Layout variety softens the collapse far more for PTv3 (0.27 vs.\ 0.70--0.72 on test2) than for PointNet++ (0.20 vs.\ 0.53--0.55), so failure severity depends on both architecture and training data. The classes that vanish also change with the regime (Table~\ref{tab:perclass}): beams and stairs under fixed layouts (109 of 180 beams and 90 of 108 stairs for PointNet++ on test2), walls and columns under randomized ones (116 of 151 walls and 115 of 160 columns on test3).

\textbf{Errors are substitutions into one sink class.} The collapse is not diffuse error. In fixed layouts, the PointNet++ baseline predicts \emph{column} on 66\% of all test2 points and 70\% of test3 points, against true shares of 12\% and 15\% (Fig.~\ref{fig:shares}a,b); the PTv3 baseline does the same on test3 (54\% vs.\ 17\%). Auditing individual failures gives a clean regularity: when a cut removes the wall behind a column, the column's points are labelled wall (188 of 188 points in the extreme case); when it removes the floor region, the same column becomes stair even where the stair itself has been cut away; beams are taken apart into column and slab. With randomized layouts the substitution survives but moves (Fig.~\ref{fig:shares}c,d): PointNet++ assigns 26.5--31.5\% of test3 points to \emph{stair} across the three seeds, a class that holds 5.9\% of them, and 58--60\% of the misclassified column and wall points end up there (first seed). The randomized PTv3 baseline has no strong sink: its largest over-prediction is only 1.2--1.8$\times$ (column or stair on test2, depending on the seed; beam on test3). In both families the sink is among the rarest training classes, which inverse-frequency weighting makes cheapest to over-predict; we state this as a hypothesis we have not tested.

\textbf{GPE repairs both failure directions.} With GPE, test2/test3 mIoU reaches 0.54--0.92 and whole-object losses fall by 64--100\% in all sixteen runs (four fixed-layout model--test pairs and twelve randomized-layout model--test--seed triples; Table~\ref{tab:main}). The gain dwarfs the seed spread. Per-seed ranges, baseline\,/\,GPE: PointNet++ mIoU .53--.55\,/\,.88--.90 (test2) and .36\,/\,.79--.85 (test3), losses 181--190\,/\,8--24 and 236--238\,/\,7--10; PTv3 mIoU .70--.72\,/\,.91--.92 and .54--.57\,/\,.89--.90, losses 18--25\,/\,2--6 and 22--30\,/\,8--9. Wall losses go to zero in every run except fixed PTv3 test3 (33$\to$1). Column losses remain in the randomized family (1--22 on test2 and 3--9 on test3 for PointNet++), which Sec.~\ref{sec:why} traces to one channel, and stairs are the least stable GPE class across seeds (1, 3 and 7 losses on PointNet++ test2). GPE also restores the class shares (Fig.~\ref{fig:shares}, blue bars).

\textbf{What this means for a robot.} Progress monitoring asks which planned columns stand. We score this check directly from the per-point predictions (Table~\ref{tab:check}). A visible column (${\geq}30$ points) is \emph{missed} if fewer than half of its points are labelled column. A \emph{phantom alarm} is raised when column-labelled points more than 0.5\,m from every true column form a vertical member, i.e., a 0.25\,m plan cell whose column points span at least 1\,m in height, i.e., what a column fitter would report as an extra column. The failure type follows the sink: fixed-layout baselines raise phantom alarms in 33--44\% of views, while randomized-layout PointNet++ misses 51--82\% of visible columns. With GPE, correctly checked views rise from 17--58\% to 71--92\% on average, and randomized-layout alarms fall to at most two views per run. For PTv3, which sees a 4,096-point subsample, the thresholds are 10 points per column and 2 per cell.

\section{Why It Happens: Cues That Are Stable in Training}
\label{sec:why}

To see which evidence the GPE model uses, we shuffle one channel group at a time within a view and measure the drop in per-view recall, using one architecture (PointNet++) and one procedure for all three datasets (Fig.~\ref{fig:perm}). We also count \emph{deletions}, views in which shuffling a group drives a class from recall ${>}0.1$ to ${\leq}0.1$.

\textbf{Fixed layouts: position carries everything.} \gle{} is the top carrier of all five classes (+0.37 to +0.44). Stairs also lean on the density channels (+0.25): with a fixed number of points per object, local density identifies the small stair plates. A random forest given only the three DPE channels separates stairs with F1 0.89 in the fixed family, but only 0.38 once sampling is area-proportional.

\textbf{Randomized layouts: shape replaces position where position varies.} Slabs and beams sit at the same height in any building and keep relying on \gle{} (+0.49, +0.57; shuffling it deletes 8 beams). Walls, whose position now varies, switch almost entirely to the overlap and planarity channels (OD +0.52, \gle{} +0.03); shuffling OD deletes 57 of 206 walls. Stairs switch to verticality (VERT +0.32, 17 of 115 deleted), and their density reliance falls to +0.02. The baseline has no shape channels to switch to, which is why it fails on exactly these classes (Table~\ref{tab:perclass}).

\textbf{Real rooms show the same split.} On S3DIS, ceiling and floor rely on \gle{} (+0.71, +0.74); walls on OD (+0.74, 125 of 139 deleted); columns on VERT (+0.75, 69 of 77 deleted). Beams, the class whose recall GPE fails to protect on real scans (Sec.~\ref{sec:real}), have the weakest carrier (\gle{} +0.30).

\textbf{The remaining failure.} Overall the randomized-layout column has no strong carrier (\gle{} +0.19). VERT helps it in six of eleven combinations (+0.14 to +0.35) but hurts under bottom+middle ($-0.47$) and middle+bottom+right ($-0.29$), the two combinations where GPE column recall stays near 0.30. Here shuffling the channel that rescues columns elsewhere \emph{raises} their recall, so the residual failure is carried by VERT itself; why the short fragments left by these cuts mislead it is open.

We draw one conclusion from the three panels: the network relies on whichever cue is stable in its training data. If position identifies a class, the model uses position, and an occlusion that moves the scene frame or removes the context that position refers to produces a confident wrong label. Supplying intrinsic shape helps exactly where training data stops making position reliable.

\section{Is Augmentation Enough?}

\begin{table}[t]
\centering
\caption{Augmentation control (PointNet++). Baseline / +augmentation / +GPE. Fixed-layout augmentation: two training seeds. One joint evaluation run, so counts differ from Table~\ref{tab:perclass} by a few views.}
\label{tab:aug}
\scriptsize
\setlength{\tabcolsep}{2.5pt}
\begin{tabular}{lcccc}
\toprule
setting & mIoU & objects lost & columns lost & walls lost\\
\midrule
fixed, test2 & .20 / .76--.81 / .78 & 271 / 14--19 / 3 & 18 / 10--14 / 1 & 50 / 0--4 / 0\\
fixed, test3 & .15 / .62--.73 / .65 & 202 / 25--35 / 0 & 15 / 14--16 / 0 & 43 / 2--6 / 0\\
rand., test2 & .55 / .69 / .88 & 185 / 63 / 24 & 80 / 22 / 21 & 93 / 24 / 0\\
rand., test3 & .36 / .65 / .79 & 235 / 68 / 10 & 115 / 24 / 9 & 115 / 31 / 0\\
\bottomrule
\end{tabular}
\end{table}

\begin{figure}[t]
\centering
\includegraphics[width=\columnwidth]{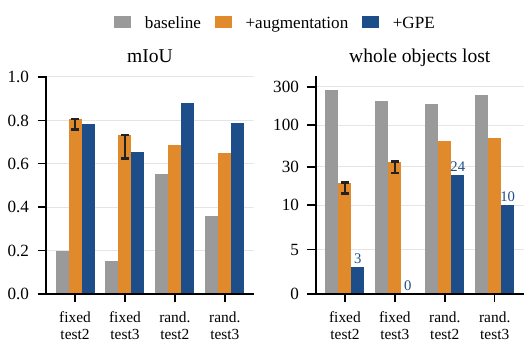}
\caption{Augmentation versus GPE (PointNet++). Whiskers: the two training seeds of the fixed-layout augmentation model. On mIoU (left) augmentation can match or beat GPE depending on the seed and regime; on whole-object losses (right, symlog scale) GPE is lower in every setting.}
\label{fig:aug}
\end{figure}

\begin{table}[t]
\centering
\caption{S3DIS Area 1 (PointNet++). IoU, baseline$\to$+GPE, and whole-object losses with exact McNemar $p$.}
\label{tab:s3dis}
\footnotesize
\begin{tabular}{lcccc}
\toprule
class & IoU & views & objects lost & $p$\\
\midrule
ceiling & .80$\to$.87 & 139 & 0$\to$0 & --\\
floor & .74$\to$.90 & 104 & 0$\to$0 & --\\
wall & .33$\to$.55 & 139 & 10$\to$0 & .002\\
beam & .40$\to$.40 & 136 & 27$\to$23 & .50 (n.s.)\\
column & .09$\to$.17 & 77 & 18$\to$8 & .064\\
\midrule
mIoU & .47$\to$.58 & & & \\
\bottomrule
\end{tabular}
\end{table}

If the model fails on an occlusion type it never saw, why not put that type into training? We rebuilt each training set at the same size (710 views): half follow the original protocol, half use scene-level two-direction cuts over all 18 non-opposing direction pairs, the machinery that generates the test sets. The augmented PointNet++ baseline is thus trained on the occlusion family of test2 (except its two three-direction combinations), while the GPE model has only ever seen column cuts. test3 is unseen for all three models. The fixed-family augmented model was trained twice, which exposed its seed variance.

\textbf{Augmentation is a strong competitor on mIoU.} It lifts mIoU in every setting and generalizes from two- to three-direction cuts (Table~\ref{tab:aug}, Fig.~\ref{fig:aug}). With fixed layouts, the two training runs reached 0.76 and 0.81 on test2 and 0.62 and 0.73 on test3, so on mIoU augmentation ties or beats GPE (0.78, 0.65) depending on the seed. With randomized layouts GPE (three seeds: 0.88--0.90 on test2, 0.79--0.85 on test3) is ahead of augmentation (one seed: 0.69, 0.65), but given the 0.05--0.11 spread we saw between two augmentation runs, we do not rest any claim on this mIoU gap.

\textbf{The tail does not move with the seed.} In all four settings and both seeds, the augmented model loses more whole objects than GPE: 14--19 vs.\ 3 and 25--35 vs.\ 0 with fixed layouts, 63 vs.\ 8--24 and 68 vs.\ 7--10 with randomized ones (GPE over its three seeds). The losses sit in different classes per regime. With fixed layouts augmentation does not help the column (18$\to$14 on test2, 15$\to$14 on test3, seed 1), while GPE takes it to 1 and 0. With randomized layouts it rescues columns (80$\to$22) but erodes walls in the very combinations it was trained on (wall recall falls by 0.17--0.35 in all four control combinations) and adds beam losses; against GPE it loses walls 24$\to$0 \mbox{($p{=}1.2$e-7)} and 31$\to$0 \mbox{($p{=}9.3$e-10)}, beams 12$\to$2 ($p{=}.0064$) and 9$\to$0 ($p{=}.0039$), and columns 22 vs.\ 21 (tie) and 24$\to$9 ($p{=}.0026$).

\textbf{The two are complementary.} Augmentation keeps column recall at 0.91--0.98 in top-containing cuts where GPE drops to 0.55--0.70 (e.g., left+back+top 0.98 vs.\ 0.55), while GPE is far ahead on walls and beams. Augmentation also presumes that the occlusion family is known in advance; on a site, the family changes as the building goes up, and simulated occlusion training depends on how representative the simulated patterns are \cite{meppnext,noichl}. Training GPE with augmentation is the natural next experiment.

\section{Does It Hold on Real Scans?}
\label{sec:real}

We test transfer on S3DIS \cite{s3dis}, remapped to six classes (ceiling, floor, wall, beam, column, other). Beams are nearly absent from the conventional Area-5 split (1.3k beam points vs.\ 73.5k in Area 1), so we choose the test area by target-class presence: test on Area 1, train on Areas 2--6. PointNet++ baseline and GPE models are retrained on real rooms; the test set holds 35 intact rooms and 104 composite-cut views produced with the synthetic protocol and the same measurability rule.

The collapse-and-repair structure transfers (Table~\ref{tab:s3dis}). Under back+middle cuts the baseline column recall falls from 0.76 to 0.18 while GPE holds 0.95$\to$0.89 (Wilcoxon, Bonferroni-adjusted \mbox{$p{=}6$e-5}), and the baseline over-predicts column as on synthetic scenes (precision .09). Walls are the strongest real-scan result: whole-wall losses fall from 10 to 0 ($p{=}.002$), with significant gains in every stratum including intact rooms. Column losses fall from 18 to 8, directionally consistent but marginal ($p{=}.064$). Beam protection does not reach significance ($p{=}.50$), matching its weak carrier in Fig.~\ref{fig:perm}. Overall mIoU improves from 0.47 to 0.58.

\section{Limitations}

\textbf{Synthetic geometry.} Both scene families are boxes sampled on visible faces, not ray-cast scans; real sites add clutter, sensor noise and view-dependent density. S3DIS tests transfer for one architecture and supports the wall and column results, not the beam result.

\textbf{Density in the fixed family.} The fixed family samples a fixed number of points per object, so density partly identifies stairs there. The randomized family removes this cue, and every main claim is reported on both families.

\textbf{Remaining GPE failures.} Columns under bottom+middle cuts stay near 0.30 recall in the randomized family, and objects cut to a few dozen points are not recovered by any variant: their surviving geometry genuinely resembles the confused class.

\textbf{Statistics.} Randomized-layout models are trained with three seeds, while fixed-layout and S3DIS models are trained once and the augmented model twice (fixed) or once (randomized); McNemar and Wilcoxon tests are over views, not training runs. The two augmentation seeds show that mIoU differences of 0.05--0.11 can arise from the seed alone, which is why we base claims on whole-object losses. S3DIS test views share rooms and are not independent. The column check of Table~\ref{tab:check} is computed from predictions with fixed thresholds; it is a proxy for a progress-monitoring pipeline, not a field test.

\textbf{Scope.} The PTv3-style model is a simplified single-resolution transformer, so its numbers describe the design family rather than the reference implementation. We approximate the robot-level consequence with an offline column check but run no robot; how such errors propagate into localization or progress reports is argued, not measured.

\section{Conclusion}

Under unseen scene-level occlusion, point-cloud segmentation does not fade. It confidently re-identifies hidden elements, and the errors collapse onto a sink class whose identity is set by the training data: the column with fixed layouts, the stair with randomized ones. Networks lean on position wherever position identified a class in training, and an occlusion that moves or removes the context that position refers to turns that reliance into a confident wrong label. Fourteen object-intrinsic geometric features remove most whole-object losses in both regimes, two architectures and three training seeds, make an as-built column check correct in most views, and on real scans remove every lost wall. Occlusion augmentation matches them on average accuracy but not on the tail. For a construction robot, a column that is occasionally invisible is a different system from one that is always seen, even at equal mIoU; occlusion studies should report whole-object losses and the class they flow into, alongside the headline score.

\section*{Acknowledgment}

Parts of the text were drafted and edited with a large language model; all experiments, analyses and claims were designed, run and verified by the author, per the IEEE-RAS policy on AI-generated content.

\end{document}